\documentclass[10pt,twocolumn,letterpaper]{article}

\usepackage[pagenumbers]{cvpr} 

\usepackage{graphicx}
\usepackage{amsmath}
\usepackage{amssymb}
\usepackage{booktabs}
\usepackage{array}
\usepackage{multirow}

\usepackage[pagebackref,breaklinks,colorlinks]{hyperref}

\usepackage{algorithm}
\usepackage{algpseudocode}
\usepackage[table]{xcolor}
\usepackage{xcolor}         

\usepackage[capitalize]{cleveref}
\crefname{section}{Sec.}{Secs.}
\Crefname{section}{Section}{Sections}
\Crefname{table}{Table}{Tables}
\crefname{table}{Tab.}{Tabs.}

\newcommand{\RR}{\mathbb{R}}

\newcommand{\cL}{\mathcal{L}}

\newcommand{\cX}{\mathcal{X}}

\newcommand{\vecb}{\mathbf{b}}

\newcommand{\vech}{\mathbf{h}}

\newcommand{\vecv}{\mathbf{v}}
\newcommand{\vecx}{\mathbf{x}}
\newcommand{\vecy}{\mathbf{y}}
\newcommand{\vecz}{\mathbf{z}}

\newcommand{\bU}{\mathbf{U}}
\newcommand{\bV}{\mathbf{V}}
\newcommand{\bW}{\mathbf{W}}

\def\cvprPaperID{7644} 
\def\confName{CVPR}
\def\confYear{2023}

\begin{document}
\title{Generalizable Implicit Neural Representations via Instance Pattern Composers}

\author{Chiheon Kim\thanks{Equal contribution} \\
Kakao Brain\\
{\tt\small chiheon.kim@kakaobrain.com}
\and
Doyup Lee\footnotemark[1] \\
Kakao Brain\\
{\tt\small doyup.lee@kakaobrain.com}
\and
Saehoon Kim\\
Kakao Brain\\
{\tt\small shkim@kakaobrain.com}
\and
Minsu Cho\\
POSTECH\\
{\tt\small mscho@postech.ac.kr}
\and
Wook-Shin Han\thanks{Corresponding author}\\
POSTECH\\
{\tt\small wshan@postech.ac.kr}
}

\maketitle



\begin{abstract}
Despite recent advances in implicit neural representations (INRs), it remains challenging for a coordinate-based multi-layer perceptron (MLP) of INRs to learn a common representation across data instances and generalize it for unseen instances.
In this work, we introduce a simple yet effective framework for generalizable INRs that enables a coordinate-based MLP to represent complex data instances by modulating only a small set of weights in an early MLP layer as an instance pattern composer; 
the remaining MLP weights learn pattern composition rules for common representations across instances.
Our generalizable INR framework is fully compatible with existing meta-learning and hypernetworks in learning to predict the modulated weight for unseen instances.
Extensive experiments demonstrate that our method achieves high performance on a wide range of domains such as an audio, image, and 3D object, while the ablation study validates our weight modulation.
\end{abstract}

\section{Introduction}

Implicit neural representations (INR) have shown the potential to represent complex data as continuous functions.
Assuming that a data instance comprises the pairs of a coordinate and its output features, INRs adopt a parameterized neural network as a mapping function from an input coordinate into its output features.
For example, a coordinate-based MLP~\cite{cppn} predicts RGB values at each 2D coordinate as an INR of an image.
Despite the popularity of INRs, a trained MLP cannot be generalized to represent other instances, since each MLP learns to memorize each data instance.
Thus, INRs necessitate separate training of MLPs to represent a lot of data instances as continuous functions.

Generalizable INRs aim to learn common representations of a MLP across instances, while modulating features or weights of the coordinate-based MLP to adapt unseen data instances~\cite{learnit,functa,transinr}.
The feature-modulation method exploits the latent vector of an instance to condition the activations in MLP layers through concatenation~\cite{deepsdf} or affine-transform~\cite{film,coin++}.
Despite the computational efficiency of feature-modulation, the modulated INRs have unsatisfactory results to represent complex data due to their limited modulation capacity.
On the other hand, the weight-modulation method learns to update the whole MLP weights to increase the modulation capacity for high performance.
However, modulating whole MLP weights leads to unstable and expensive training~\cite{maml,learnit,functa,transinr}.

\begin{figure}
    \centering
    \includegraphics[height=0.45\textwidth, width=0.45\textwidth]{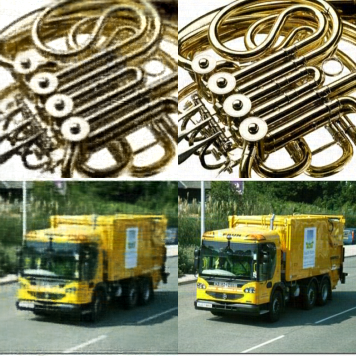}
    \caption{The reconstructed images of 178$\times$178 ImageNette by TransINR~\cite{transinr} (left) and our generalizable INRs (right).}
    \label{fig:teaser}
    \vspace{-0.1in}
\end{figure}

In this study, we propose a simple yet effective framework for generalizable INRs via \emph{Instance Pattern Composers} to modulate only a small set of MLP weights.
We postulate that a complex data instance can be represented by composing low-level patterns in the instance~\cite{cppn}.
Thus, we rethink and categorize the weights of MLP into i) \emph{instance pattern composers} and ii) \emph{pattern composition rule}.
The instance pattern composer is a weight matrix in the early layer of our coordinate-based MLP to extract the instance content patterns of each data instance as a low-level feature.
The remaining weights of MLP is defined as a pattern composition rule, which composes the instance content patterns in an instance-agnostic manner.
In addition, our framework can adopt both optimization-based meta-learning and hypernetworks to predict the instance pattern composer for an INR of unseen instance.
In experiments, we demonstrate the effectiveness of our generalizable INRs via instance pattern composers on various domains and tasks.

Our main contributions are summarized as follows.
1) \emph{Instance pattern composers} enable a coordinate-based MLP to represent complex data by modulating only one weight, while \emph{pattern composition rule} learns the common representation across data instances.
2) Our instance pattern composers are compatible with  optimization-based meta-learning and hypernetwork to predict modulated wights of unseen data during training.
3) We conduct extensive experiments to demonstrate the effectiveness of our framework through quantitative and qualitative analysis.

\section{Related Work}
\label{sec:related_work}
\paragraph{Implicit neural representations (INRs).}
INRs train a parameterized neural network to represent complex and continuous data such as audios, images, and 3D objects and scenes.
Seminal works incorporate Fourier features~\cite{nerf,Zhong2020Reconstructing,ffnet} and sinusoidal activations~\cite{siren} in the coordinate-based MLP to avoid the spectral bias~\cite{rahaman2019spectral,basri2020frequency}.
Consequently, recent advances of INRs have shown broad impacts on various applications such as data reconstruction and compression~\cite{coin,siren,ffnet,nerv}, and 3D representations~\cite{lu2021compressive,siren,ffnet,nerf,lightfield,mipnerf}.
A coordinate-based MLP can learn to represent each data instance with high-resolution and complex patterns, but the learned MLP cannot be generalized to represent other data instances and requires re-training from the scratch.

\paragraph{Generalizable INRs.}
Generalizable INRs learn to modulate or adapt the coordinate-based MLP to unseen data instances.
Given a latent vector of each data instance, autodecoding~\cite{occupancynet,deepsdf} concatenates the latent vector into the features of MLP as the input condition, while sharing whole MLP weights across data instances.
Inspired by the success of feature modulations~\cite{film,StyleGANv2}, a hypernetwork~\cite{hypernetworks} is trained to predict the modulation vectors for each data instance to scale and shift the activations in all layers of the shared MLP~\cite{coin++,functa,modulated_liif}.
Both approaches are simple and computationally efficient, since they do not need to change the whole weights of MLP.
However, the scope and capacity of feature modulations are limited and insufficient to adapt the shared MLP for a multitude of data instances.

Existing studies adopt optimization-based meta-learning to training generalizable INRs.
The bilevel optimizations such as MAML~\cite{maml} and CAVIA~\cite{cavia} train the weight initialization of coordinate-based MLP, where the inner optimization achieves rapid adaptation of MLP to unseen data instances in a few gradient steps~\cite{metasdf,learnit}.
Despite high performance using direct weight updates, the training is unstable and memory intensive due to the computation of high-order gradients~\cite{functa} and requires an exhaustive search of hyperparameters.
Interpreting the inner optimization of MAML as the inference of transformers~\cite{transformer,ViT}, TransINR~\cite{transinr} uses a hypernetwork comprised of a transformer to predict the column vectors in the weight matrix at every MLP layer.

\begin{figure*}
    \centering
    \includegraphics[width=\textwidth]{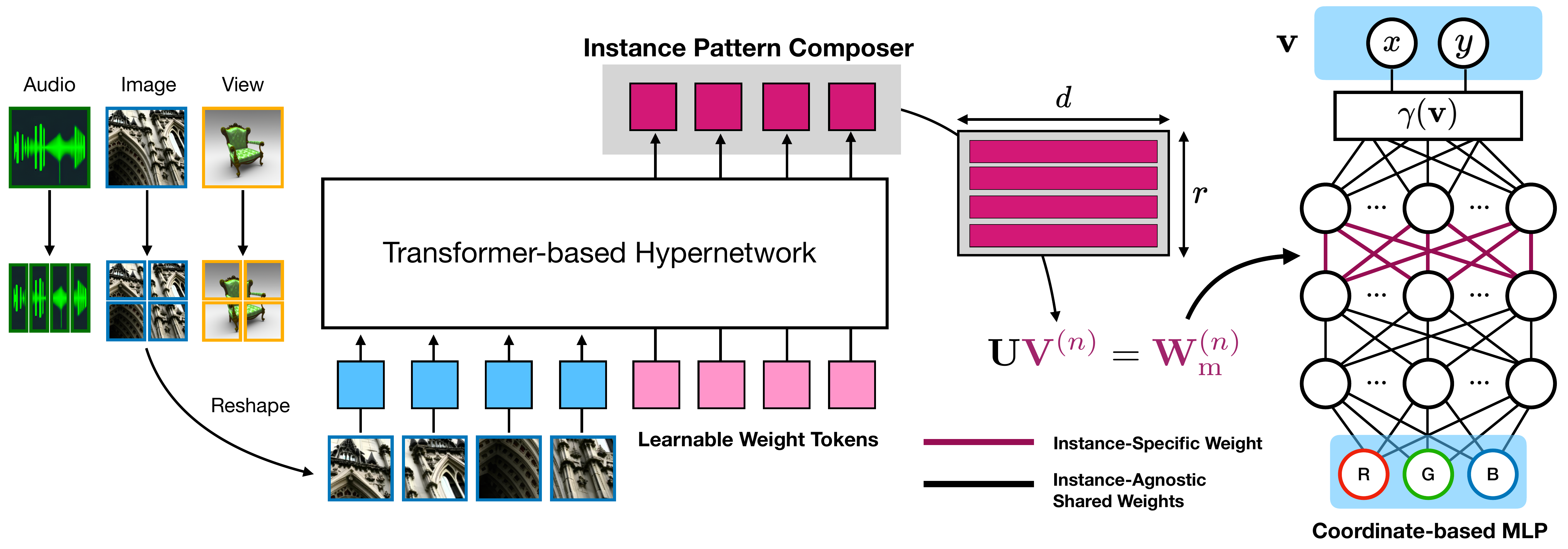}
    \caption{Overview of our framework with Instance Pattern Composer for generalizable INRs. Instance pattern composer modulates the weight matrix in the second lowest MLP layer, while the remaining weights learns an instance-agnostic pattern composition rule.}
    \label{fig:framework}
    \vspace{-0.1in}
\end{figure*}

\section{Methods}
\label{sec:method}

In this section, we propose an effective framework for generalizable INRs via instance pattern composer. 
We first present the formulation of INR for a data instance and generalize it for multiple instances in a dataset.
Then, we propose the \emph{instance pattern composer} for our generalizable INRs to modulate a small set of weights in the second MLP layer, and \emph{pattern composition rule} to generate complex data based on the extracted patterns.
Finally, we explain how our framework can be combined with optimization-based meta-learning and transformer-based hypernetwork to preidct the instance pattern composer of data instances.

\subsection{Preliminary}
An INR represents a data instance as a continuous function by learning a parameterized neural network, \eg, a coordinate-based MLP, which maps a coordinate into its corresponding features.
Given a dataset $\cX = \{\vecx^{(n)}\}_{n=1}^N$ with $N$ instances, we assume that each instance $\vecx^{(n)}$ is expressed by $M_n$ pairs of an input coordinate $\vecv^{(n)}_{i}$ and its corresponding output feature $\vecy^{(n)}_i$: 
\begin{equation}
\label{eq:data_example}
\vecx^{(n)}=\{ (\vecv^{(n)}_{i}, \vecy^{(n)}_{i}) \}_{i=1}^{M_n},
\end{equation}
where $\vecv^{(n)}_i \in \RR^{d_\text{in}}$  and $\vecy^{(n)}_i \in \RR^{d_\text{out}}$.
For example, an $H \times W$ image $\vecx^{(n)}$ takes as input a 2D coordinate ($d_\text{in}=2)$ and produces as output its RGB values ($d_\text{out}=3$), resulting in $M_n=H W$ for all pixels. 
Given trainable parameters $\phi^{(n)}$, a coordinate-based MLP $f_{\phi^{(n)}}: \RR^{d_\text{in}} \to \RR^{d_\text{out}}$ learns to implicitly represent an instance $\vecx^{(n)}$, while minimizing the mean-squared error over coordinates:
\begin{equation}
\label{eq:inr_loss}
    \cL_n(\phi^{(n)};\vecx^{(n)}) := \frac{1}{M_n}\sum_{i=1}^{M_n} \| \vecy^{(n)}_i - f_{\phi^{(n)}} (\vecv^{(n)}_i) \|_2^2.
\end{equation}
Note that since a coordinate-based MLP model $f_{\phi^{(n)}}$ is trained to represent only \emph{one} instance $\vecx^{(n)}$, the model cannot represent other instances in the dataset as well as unseen instances after training.
The straightforward approach to obtain INRs for a dataset is to separately train an MLP per instance from scratch, but it is computationally infeasible for a large-scale dataset.
Furthermore, this separate training cannot leverage common information across instances, limiting efficiency and generalizability. 

\subsection{Generalizable Implicit Neural Representations with Instance Pattern Composers}
To overcome the limitation of conventional INRs, we propose the framework for generalizable INRs with \emph{Instance Pattern Composers}.
Our framework can efficiently modulate a small set of MLP weights, while learning common representations across instances.
After we present the formulation of generalizable INRs, we explain the details of our framework with instance pattern composers.

\subsubsection{Generalizable INRs.} 
We define the two types of parameters of a coordinate-based MLP for generalizable INRs: i) instance-specific parameter $\phi^{(n)}$ for an instance $\vecx^{(n)}$ and ii) instance-agnostic parameter $\theta$.
The instance-specific parameter $\phi^{(n)}$ characterizes a data instance $\vecx^{(n)}$, while the instance-agnostic parameter $\theta$ is shared across all instances to learn the underlying structural information in a dataset.
Then, the coordinate-based MLP for generalizable INRs is trained to minimize the average of mean-square errors over training dataset $\cX$: 
\begin{equation}
\label{eq:ginr_loss}
    \cL(\theta, \{\phi^{(n)}\}_{n=1}^N; \cX) := \frac{1}{N} \sum_{n=1}^{N} \cL_n(\theta, \phi^{(n)}; \vecx^{(n)}).
\end{equation}
Previous studies first train whole MLP weights as the instance-agnostic parameter $\theta$, and then a modulation function $g$ is used to convert the whole MLP weights to be instance-specific as $\phi^{(n)} = g(\theta, \vecx^{(n)})$, where $g$ is executing the inner optimization steps in meta-learning~\cite{maml,learnit} or directly predicting weights in hypernetwork~\cite{hypernetworks,transinr}.
However, we remark that modulating whole MLP weights is limited in terms of efficiency and effectiveness due to unstable training and expensive computational costs.

\subsubsection{Generalizable INRs by Modulating One Weight Matrix as Instance Pattern Composer}
We propose a simple yet powerful weight modulation of coordinate-based MLPs for generalizable INRs.
Inspired by the first usage of coordinate-based MLP, which composes low-level patterns to synthesize complex visual patterns~\cite{cppn}, we postulate that our coordinate-based MLP can represent complex data by composing simple patterns of instance-specific contents.
Thus, we first categorize the weights of MLP into the following two types: i) \emph{Instance Pattern Composer} as instance-specific parameter $\phi^{(n)}$, and ii) \emph{Pattern Composition Rule} as instance-agnostic parameter $\theta$.
Especially, we assign one weight matrix in the early MLP layer for the instance pattern composer to be modulated, while the remaining weights are instance-agnostic pattern composition rule.
For the brevity of notation, we omit the subscript $i$ and denote $\vecv^{(n)}_i$ as $\vecv^{(n)}$ in this section.

\paragraph{Low-level frequency patterns.}
We convert a coordinate $\vecv^{(n)}$ of an instance $\vecx^{(n)}$ into its Fourier features $\gamma(\vecv^{(n)}) \in \RR^{d_\text{f}}$ with a dimensionality $d_\text{f}$~\cite{ffnet}.
Then, a fully-connected (FC) layer generates low-level frequency patterns $\vech_\mathrm{f}$ as 
\begin{equation} \label{eq:ff}
   \vech_\mathrm{f} = \sigma(\bW_\mathrm{f} \gamma(\vecv^{(n)}) + \vecb_\mathrm{f}),
\end{equation}
where $\bW_\mathrm{f} \in \RR^{d \times d_\text{f}}$ and $\vecb_\mathrm{f} \in \RR^d$ are a learnable weight matrix and a bias vector, respectively, $d$ is the dimensionality of hidden layers, and $\sigma (\cdot)$ is an element-wise nonlinearity function \eg ReLU.
Since $\bW_\mathrm{f}$ and $\vecb_\mathrm{f}$ are instance-agnostic, data instances have the same frequency patterns $\vech_\mathrm{f}$.


\paragraph{Instance Pattern Composer.} 
An \emph{instance pattern composer} characterizes the INR of a data instance $\vecx^{(n)}$ and extracts \emph{instance content patterns} $\vech^{(n)}$ based on frequency patterns $\vech_\mathrm{f}$.
Given a modulated weight matrix $\bW_\mathrm{m}^{(n)} \in \RR^{d \times d}$, we define an instance pattern composer $\bV^{(n)} \in \RR^{r \times d}$ of an instance $\vecx^{(n)}$ as a factorized matrix with rank $r$
\begin{equation} \label{eq:W=UV}
    \bW_\mathrm{m}^{(n)} = \mathbf{U} \mathbf{V}^{(n)},
\end{equation}
where $\bU \in \RR^{d \times r}$ is an instance-agnostic weight.
Then, an instance content pattern $\vech^{(n)}$ of $\vecx^{(n)}$ is predicted as
\begin{equation}
    \vech^{(n)} = \sigma(\bW_\mathrm{m}^{(n)} \vech_\mathrm{f} + \vecb_\mathrm{m}),
\end{equation}
where $\vecb_\mathrm{m} \in \RR^{d}$ is an instance-agnostic bias vector.
Note that the instance pattern composer $\bV^{(n)}$ extracts the instance-specific representations of $\vecx^{(n)}$ to characterize an instance of modulated MLPs as a continuous representation of $\vecx^{(n)}.$
Thus, our generalizable INRs only modulate the one weight matrix $\bV^{(n)}$ to represent complex data, while sharing other MLP weights across data instances.

\paragraph{Pattern composition rule.}
Based on the instance content patterns $\vech^{(n)}$ at coordinate $\vecv^{(n)}$, the subsequent FC layers are trained to predict the output features $\vecy^{(n)}$.
We assume that the subsequent MLP layers learn to compose the instance content patterns $\vech^{(n)}$ to represent complex output features $\vecy^{(n)}$, while learning the underlying structural information across data instances.
Specifically, when the total number of MLP layers is $L$, the parameters of the remaining $L-2$ layers are shared across data instances to determine the \emph{pattern composition rule} of MLP.
Given $\vecz_2^{(n)} := {\vech}^{(n)}$, the remaining hidden activations of MLP are computed as 
\begin{equation}
    \vecz_\ell^{(n)} = \sigma(\bW_\ell \vecz_{\ell-1}^{(n)} + \vecb_\ell),
\end{equation}
where $\bW_\ell \in \RR^{d\times d}$ and $\vecb_\ell \in \RR^{d}$ are weight and bias of $l$-th layer, and $l \in \{3, \cdots, L-1\}$.
Finally, given the weight $\bW_L \in \RR^{d_\text{out}\times d}$ and bias $\vecb_L \in \RR^{d_\text{out}}$, the output layer predicts the output features $\vecy^{(n)}$ as 
\begin{equation}
    f_{\theta,\phi^{(n)}}(\vecv^{(n)}) := \bW_{L} \vecz_{L-1}^{(n)} + \vecb_L,
\end{equation}
where $\phi^{(n)} = \bV^{(n)}$ is the instance-specific parameter, and $\theta = \{\bW_\mathrm{f}, \vecb_\mathrm{f}, \mathbf{U}, \vecb_\mathrm{CP}, \bW_\mathrm{3}, \vecb_\mathrm{3}, \cdots, \bW_\mathrm{L-1}, \vecb_\mathrm{L-1}\}$ is the instance-agnostic parameter.
Since the pattern composition rule is the set of instance-agnostic parameters, our coordinate-based MLP shares all trainable parameters except for $\bV^{(n)}$.
That is, our generalizable INRs use the same rule to compose the content patterns $\vech^{(n)}$ of different instances to represent complex data instances.
In summary, our generalizable INRs learn the common pattern composition rule of extracted instance content patterns to generalize the learned representations for unseen data instances.

\begin{algorithm}[t]
\caption{Optimization-based meta-learning for generalizable INRs via instance pattern composer.}
\label{alg:cavia}
\begin{algorithmic}[1]
\Require Randomly initialized $\theta$, $\phi$, a dataset $\mathcal{X}$, the number of inner steps $N_\text{inner}$, and learning rates $\epsilon, \epsilon'$.
\While{not done}
\For{$n=1, \cdots, N$}
    \State Initialize instance-specific parameter $\phi^{(n)} \gets \phi$ 
\EndFor

\item[] {\color{gray} \texttt{\quad /* inner-loop updates for $\theta^{(n)}$ */}}
\For{\textbf{all} step $\in \{1, \cdots, N_\text{inner}\}$ and $\vecx^{(n)} \in \mathcal{X}$}
\State $\phi^{(n)} \gets \phi^{(n)} - \epsilon \| \phi^{(n)} \|^2 \nabla_{\phi^{(n)}} \cL_n(\theta, \phi^{(n)}; \vecx^{(n)})  $
\EndFor

\item[] {\color{gray} \texttt{\quad /* outer-loop updates for $\theta$, $\phi$ */}}
\State Update $\phi \gets \phi - \epsilon' \nabla_\phi \cL(\theta, \{\phi^{(n)}\}_{n=1}^N; \cX)$

\State Update $\theta \gets \theta - \epsilon' \nabla_\theta \cL(\theta, \{\phi^{(n)}\}_{n=1}^N; \cX)$

\EndWhile
\end{algorithmic}
\end{algorithm}

\subsection{Predicting Modulation Weights}
\label{sec:weight_prediction}
Thanks to the simple method of weight modulation, our framework for generalizable INRs is compatible with existing methods to predict the modulated weight $\phi^{(n)}=\bV^{(n)}$ to characterize the INR of $\vecx^{(n)}$.
This section shows that how optimization-based meta-learning~\cite{maml,cavia,coin++,functa,learnit} and hypernetworks~\cite{hypernetworks,transinr} can be combined with our framework.

\paragraph{Optimization-based meta-learning.}
An optimization-based meta-learning can learn the initialization of instance-specific parameter $\phi^{(n)}=\phi$ to be adapted to $\vecx^{(n)}$ in few optimization steps of Eq.~\eqref{eq:inr_loss}.
Since we do not require the adaptation of the whole weights in the test time, we modify CAVIA~\cite{cavia} for our generalizable INRs in Algorithm~\ref{alg:cavia}.
Different from the original CAVIA, we train the initialization of $\phi^{(n)}$ as $\phi$ in the outer update to encourage the training of $\bU$ in Eq.~\eqref{eq:W=UV}.
We also scale the learning rate $\epsilon$ in the inner loop by the square of the norm of adapted parameter $\| \phi^{(n)} \|^2$ to improve the stability of the inner-loop updates.

\paragraph{Transformer-based hypernetwork.}
Our framework can adopt the transformer-based hypernetwork in Figure~\ref{fig:framework} to predict the $r$ number of row vectors in Instance Pattern Composer $\bV^{(n)}$ for each instance $\vecx^{(n)}$.
Specifically, we first patchify a data instance $\vecx^{(n)}$, such as an audio, image, or multiple views, into non-overlapping patches and convert them into a sequence of data tokens in the raster-scan ordering. 
Then, we concatenate $r$ learnable tokens into the sequence of data tokens, and use the concatenated token sequence as the input of the bidirectional transformer.
Finally, $r$ output tokens corresponding to learnable query tokens are linearly mapped into $\RR^{d}$ to form an $r \times d$ matrix to predict the instance-specific factorized matrix $\bV^{(n)}$ in Eq.~\eqref{eq:W=UV}.
Since the transformer predicts the instance-specific weights $\theta^{(n)} = \bV^{(n)}$, the parameters of the transformer are trained in an end-to-end manner by the optimization process of Eq.~\eqref{eq:ginr_loss}.
Although the transformer-based hypernetwork has been already proposed, our framework does not require a heuristic method of weight grouping~\cite{transinr}, but significantly improve the performance of the hypernetwork.

\section{Experiments}
\label{sec:exp}
We evaluate our framework on a wide range of domains such as audios, images, and 3D objects.
We mainly use the transformer-based hypernetwork in Figure~\ref{fig:framework}, since its training does not requires exhaustive hyperparameter search.
Nonetheless, we also validate that our framework is also compatible with optimization-based meta-learning in Section~\ref{exp:ablation_meta}.
The implementation details are in the Appendix.


\begin{table} 
\centering
\small
\caption{PSNRs of the reconstruction of the LibriSpeech test-clean dataset whose sample is trimmed into one and three seconds.}
\label{tab:audio}
\begin{tabular}{l|cc}
\toprule 
         & LibriSpeech (1s) & LibriSpeech (3s)  \\ \hline
TransINR & 39.22 & 33.17  \\ 
Ours     & \textbf{40.11} & \textbf{35.38}  \\
\bottomrule
\end{tabular}
\end{table}

\subsection{Audio Reconstruction}
\label{exp:audio_recon}
Our framework is trained on LibriSpeech-clean~\cite{librispeech} for audio reconstruction.
The MLP has five layers with $d=256$, $d_\text{in}=1$, and $d_\text{out}=1$ for an audio.
Our transformer-based hypernetwork predicts $r=256$ weight tokens for $\bV^{(n)}$, while TransINR predicts 257 weight tokens to modulate whole MLP weights via weight grouping~\cite{transinr}.
We train our framework on randomly cropped audio during 1000 epochs, while test audio is trimmed for evaluation.

Table~\ref{tab:audio} shows the PSNRs of reconstructed audios.
Although the reconstructed audios with three seconds have lower PSNRs than one second of audios, our framework consistently outperforms TransINR.
Since the main difference from TransINR is the weight modulation method, the results validate the effectiveness of our instance pattern composers to modulate a small set of MLP weights.

\begin{table} 
\centering
\small
\caption{PSNRs of reconstructed images for178$\times$178 resolution of images in the CelebA, FFHQ, and ImageNette test dataset.}
\label{tab:image178}
\begin{tabular}{l|ccc}
\toprule 
         & CelebA & FFHQ & ImageNette \\ \hline
Learned Init~\cite{learnit} & 30.37 & - & 27.07 \\
TransINR  & 33.33 & 33.66 & 29.77 \\ \hline
Ours     & \textbf{35.93} & \textbf{37.18} & \textbf{38.46} \\
\bottomrule
\end{tabular}
\end{table}

\begin{figure}
    \centering
    \includegraphics[height=0.45\textwidth, width=0.45\textwidth]{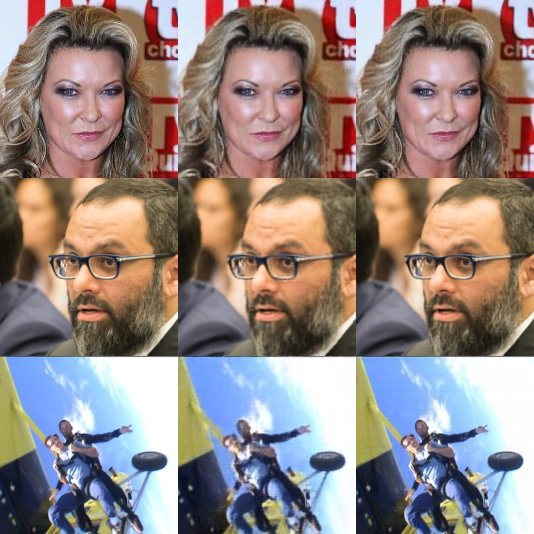}
    \caption{The reconstruction examples of TransINR~\cite{transinr} (middle) and our framework (right), given 178$\times$178 original images (left) in CelebA, FFHQ, and ImageNette in each row, respectively.}
    \label{fig:imgrec_178}
\end{figure}

\begin{figure*}
    \centering
    \includegraphics[height=0.25\textwidth, width=\textwidth]{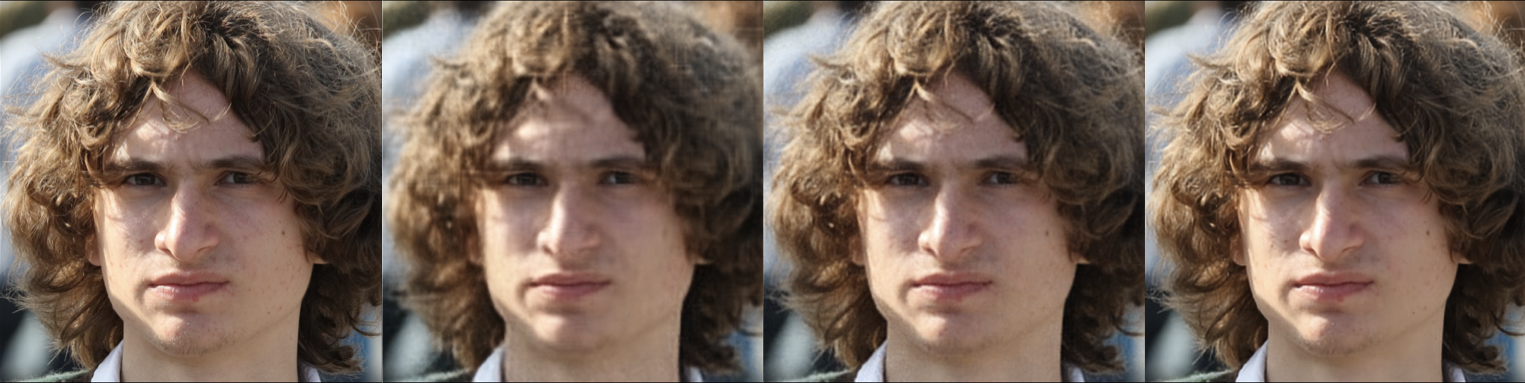}
    \caption{Examples of original 512$\times$512 images (left), and reconstructed images by TransINR~\cite{transinr} with $d=256$ and $r=259$ (middle left), our framework with $d=256$ and $r=256$ (middle right), and $d=1024$ and $r=1024$ (right).}
    \label{fig:imgrec_512}
    \vspace{-0.1in}
\end{figure*}

\begin{table} 
\centering
\small
\caption{PSNRs on high-resolution FFHQ reconstruction according to MLP dimensions $d$ and the number of weight tokens $r$.}
\label{tab:highres_imgrec}
\begin{tabular}{l|cc|cc}
\toprule 
         & $d$  & $r$            & 256$\times$256 & 512$\times$512 \\ \hline
TransINR & 256  & 64$\times$4+3  & 30.96         & 29.35 \\
TransINR & 256 & 256$\times$4+3 & 32.92        & 31.00 \\
Ours     & 256  & 256            & \textbf{34.68}         & \textbf{31.58} \\ \hline
TransINR & 1024 & 64$\times$4+3  & 33.83         & 31.57 \\
TransINR & 1024 & 256$\times$4+3 & 36.50         & 32.68 \\ 
Ours     & 1024 & 256            & 38.43         & 35.22  \\
Ours     & 1024 & 1024           & \textbf{40.37}         & \textbf{36.27} \\
\bottomrule
\end{tabular}
\end{table}

\subsection{Image Reconstruction}
\label{exp:image_recon}
We evaluate our generalizable INRs on image reconstruction of 178$\times$178, 256$\times$256, 512$\times$512 resolution of images usingfive layers of MLPs with $d_\text{out}=3$ and $d_\text{in}=2$.

\paragraph{178$\times$178 Image Reconstruction}
We evaluate our generalizable INRs of MLP with $d=256$ on 178$\times$178 image reconstruction.
Our transformer uses $r=256$ weight tokens, since TransINR uses 259 (64$\times$4+3) weight tokens to modulate all MLP layers~\cite{transinr}.
Table~\ref{tab:image178} shows that our framework significantly outperforms Learned Init~\cite{learnit} and TransINR on the three datasets by a large margin. 
TransINR cannot precisely reconstruct the images of ImageNette, which contains complex patterns in images, but our framework produces high quality of reconstructed images. 
Figure~\ref{fig:imgrec_178} shows that our framework reconstructs images with high precision.

\paragraph{High-Resolution Image Reconstruction}
We evaluate our framework on high-resolution FFHQ images with 256$\times$256 and 512$\times$512 resolutions.
As high-resolution images would require a larger capacity of INRs, MLP models with $d=256$ and $d=1024$ are modulated by instance pattern composers with $r=256$ and $r=1024$.
In Table~\ref{tab:highres_imgrec}, our framework significantly outperforms previous TransINR on high-resolution image reconstruction in various settings.
When we increase the $d$ to $1024$, our framework significantly improves PSNRs for 256$\times$256 and 512$\times$512 images.
The results show that our coordinate-based MLP can adapt to unseen data despite the minimal changes in MLP weights.
In Figure~\ref{fig:imgrec_512}, TransINR cannot reconstruct high-frequency details of original images, but our framework precisely reconstructs those details.
Considering that previous studies have not achieved high performance on high-resolution images, our results demonstrate that the weight modulation is the key to generalizable INRs.

\begin{table} 
\centering
\small
\caption{Performace comparison of generalizable INRs on novel view synthesis from a single support view.}
\label{tab:nvs}
\begin{tabular}{l|ccc}
\toprule 
                              & Chairs  & Cars   & Lamps \\ \hline
Matched Init~\cite{learnit}    & 16.30 & 22.39 & 20.79 \\
Shuffled Init~\cite{learnit}   & 10.76 & 11.30 & 13.88 \\ \hline
Learned Init~\cite{learnit}    & 18.85 & 22.80 & 22.35 \\ 
TransINR                       & 19.05 & \textbf{24.18} & 22.89 \\ \hline
Ours                           & \textbf{19.30} & \textbf{24.18} & \textbf{23.41}\\
\bottomrule
\end{tabular}
\end{table}

\begin{figure}
    \centering
    \includegraphics[height=0.24\textwidth, width=0.36\textwidth]{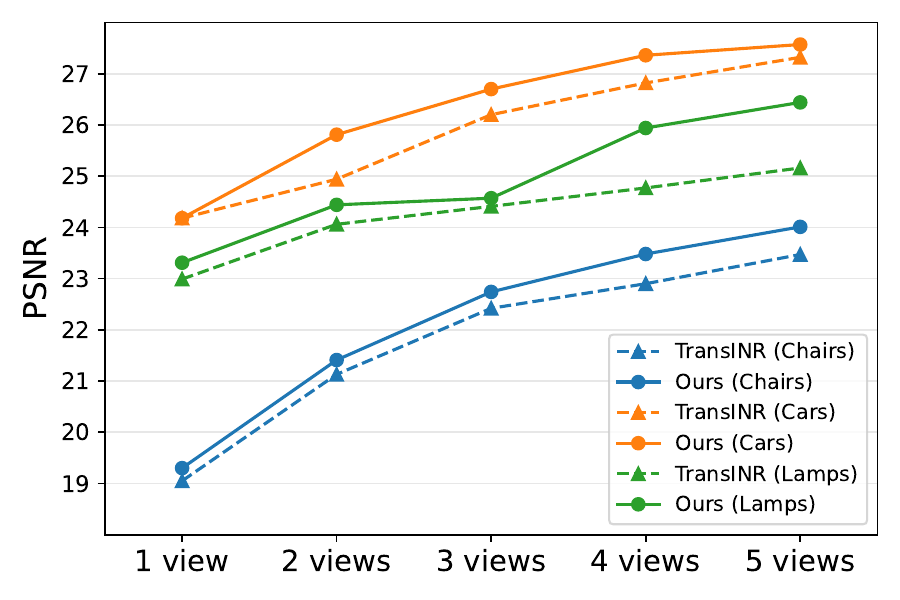}
    \caption{PSNRs on novel view synthesis of Chairs, Cars, Lamps according to the number of support views (1-5 views).}
    \label{fig:nvs_view_exp}
\end{figure}

\subsection{Novel View Synthesis}
\label{exp:nvs}
We evaluate our framework on novel view synthesis of a 3D object based on the ShapeNet Chairs, Cars, and Lamps datasets.
Given a 3D object and a few view images with known camera poses, we train the coordinate-based MLP, which has six layers with $d=256$, $d_\text{in}=3$ for ($x,y,z$) coordinates, and $d_\text{out}=4$ for outputs of RGB values and its density, to estimate the view of a 3D object under unseen camera poses.
For evaluation, we randomly sample a camera pose.
To synthesize a novel view image, we use the simple volumetric rendering~\cite{nerf} to focus on the effectiveness of our weight modulation method instead of achieving state-of-the-art performance.
We follow the experimental settings of previous studies~\cite{learnit,transinr} except for the manual decay of learning rate in TransINR~\cite{transinr}, but use a constant learning rate until the training converges.

\begin{figure}
    \centering
    \includegraphics[height=0.315\textwidth, width=0.45\textwidth]{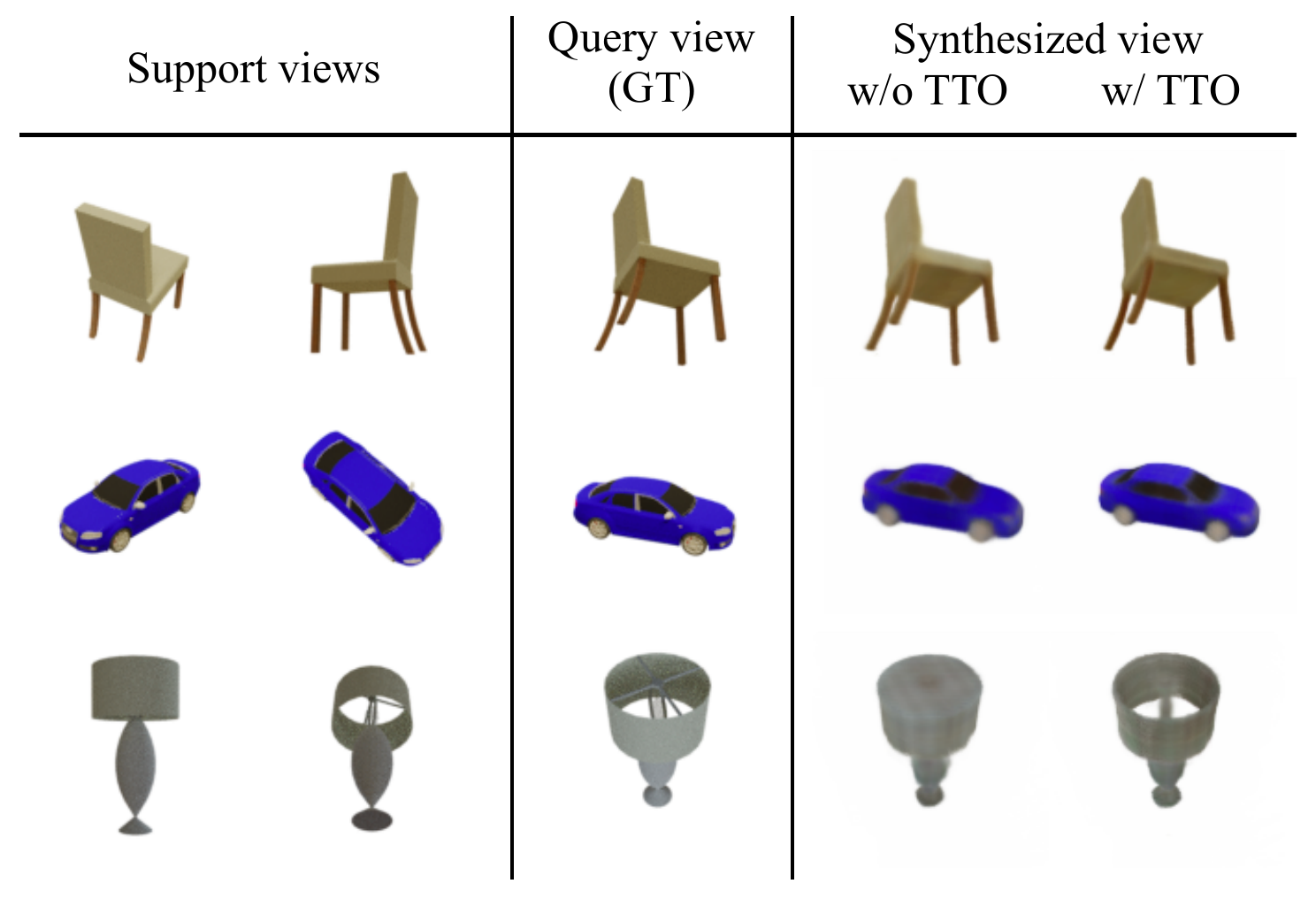}
    \caption{Novel view synthesis examples by our framework with two support views of ShapeNet Chairs, Cars, and Lamps.}
    \label{fig:nvs_example}
    \vspace{-0.1in}
\end{figure}

In Table~\ref{tab:nvs}, our generalizable INRs outperform previous approaches on novel view synthesis under a single support view.
Note that the results of our framework and TransINR are not benefited from the test-time optimization (TTO), but the other approaches use TTO by the nature of optimization-based meta-learning. 
Figure~\ref{fig:nvs_view_exp} also shows our framework consistently outperforms TransINR as the number of support views increases, while our performance is continuously improved.
Although Figure~\ref{fig:nvs_example} shows that our framework provides blurry views due to the simple volumetric rendering, our framework can capture and synthesize the shapes and colors of 3D objects based on given support views.

\begin{table} 
\centering
\small
\caption{The improvements after test-time optimization (TTO) on novel view synthesis of ShapeNet-Lamps with support views.}
\label{tab:nvs_tto}
\begin{tabular}{l|ccccc}
\toprule 
& \multicolumn{5}{c}{the number of views} \\ \hline
                    & 1   & 2  & 3  & 4  & 5  \\ \hline
TransINR            & 22.99   & 24.06   & 24.41   & 24.77   & 25.16   \\
\footnotesize{w/ TTO (all weights)}           & 25.13   & 27.28   & 28.09   & 28.56   & 28.93 \\ \hline
Ours                & 23.40   & 24.33   & 24.62   & 26.05   & 26.79  \\
\footnotesize{w/ TTO ($\bV^{(n)}$)}     & \textbf{25.47}   & 27.53   & 28.34   & \textbf{29.52}   & \textbf{30.19} \\
\footnotesize{w/ TTO (all weights)}                & 25.40   & \textbf{27.57}   & \textbf{28.40}   & 29.43   & 30.07 \\
\bottomrule
\end{tabular}
\end{table}

Table~\ref{tab:nvs_tto} shows the performance after 100 TTO steps on ShapeNet-Lamps with 1-5 support views.
We use the following two types of TTO.
The first approach optimizes the whole MLP weights, but the other only optimizes one weight matrix of instance pattern composer $\bV^{(n)}$.
Table~\ref{tab:nvs_tto} shows that our framework consistently outperforms TransINR after TTO.
Moreover, despite updating only one weight matrix, the improvement after TTO of $\bV^{(n)}$ is competitive with or even better than TTO of all weights.
In other words, our model learns a generalizable and instance-agnostic pattern composition rule to achieve high performance if the instance pattern composers $\bV^{(n)}$ can be accurately predicted.
Figure~\ref{fig:nvs_example} demonstrates that the synthesized images also become sharp and precise after TTO.


\subsection{Ablation Study}
\label{exp:ablation}

\begin{table} 
\centering
\small
\caption{PSNRs of our generalizable INRs on ImageNette and Lamps (2 views) according to the types of modulation methods.}
\label{tab:ablation_mod_type}
\begin{tabular}{l|cc}
\toprule 
$\bW_\mathrm{CP}^{(n)}$                    & ImageNette & Lamps (2 views)  \\ \hline
$\mathbf{V}^{(n)}$                         & 30.01 & \textbf{24.69} \\ 
$\mathbf{U}^{(n)} \mathbf{V}^{(n)}$        & 32.35 & 23.04 \\
$\mathbf{U} \odot \mathbf{V}^{(n)}$        & 30.64 & 24.18 \\ \hline
$\mathbf{U} \mathbf{V}^{(n)}$ (ours)       & \textbf{35.93} & 24.44 \\ 
\bottomrule
\end{tabular}
\end{table}

\begin{table} 
\centering
\small
\caption{PSNRs of our generalizable INRs on image reconstruction according to the location of modulated weights in MLP.}
\label{tab:ablation_mod_scope}
\begin{tabular}{l|ccccc}
\toprule 
& \multicolumn{5}{c}{the modulated layer of MLP} \\ \hline
& 1 & 2 & 3 & 4 & 5\\  \hline
ImageNette & 31.00 & \textbf{35.93} & 32.99 & 31.10 & 20.26 \\ 
FFHQ & 36.04 & \textbf{36.20} & 34.2 & 31.09 & 22.92 \\  
\bottomrule
\end{tabular}
\end{table}

\begin{figure}
    \centering
    \includegraphics[height=0.26\textwidth, width=0.39\textwidth]{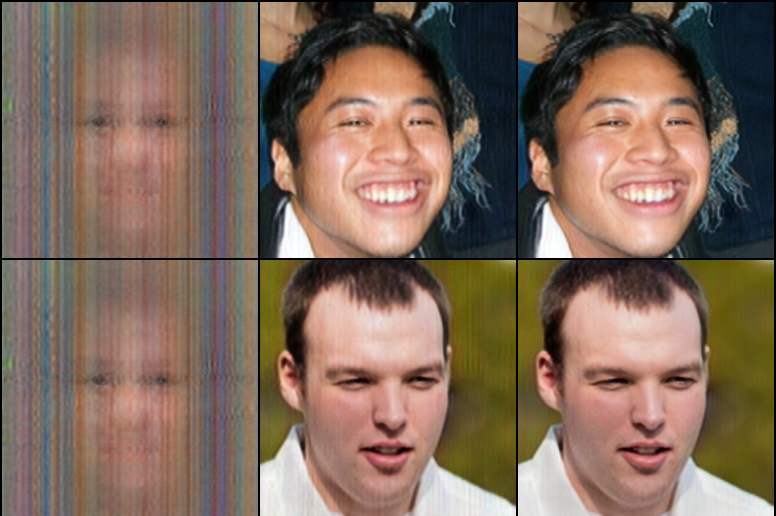}
    \caption{Reconstructions of FFHQ after inner-loop updates for $\bV^{(n)}$ (left: initial, middle: first update, right: second update).}
    \label{fig:example_cavia}
    \vspace{-0.1in}
\end{figure}

\begin{figure*}
    \centering
    \includegraphics[width=0.9\textwidth]{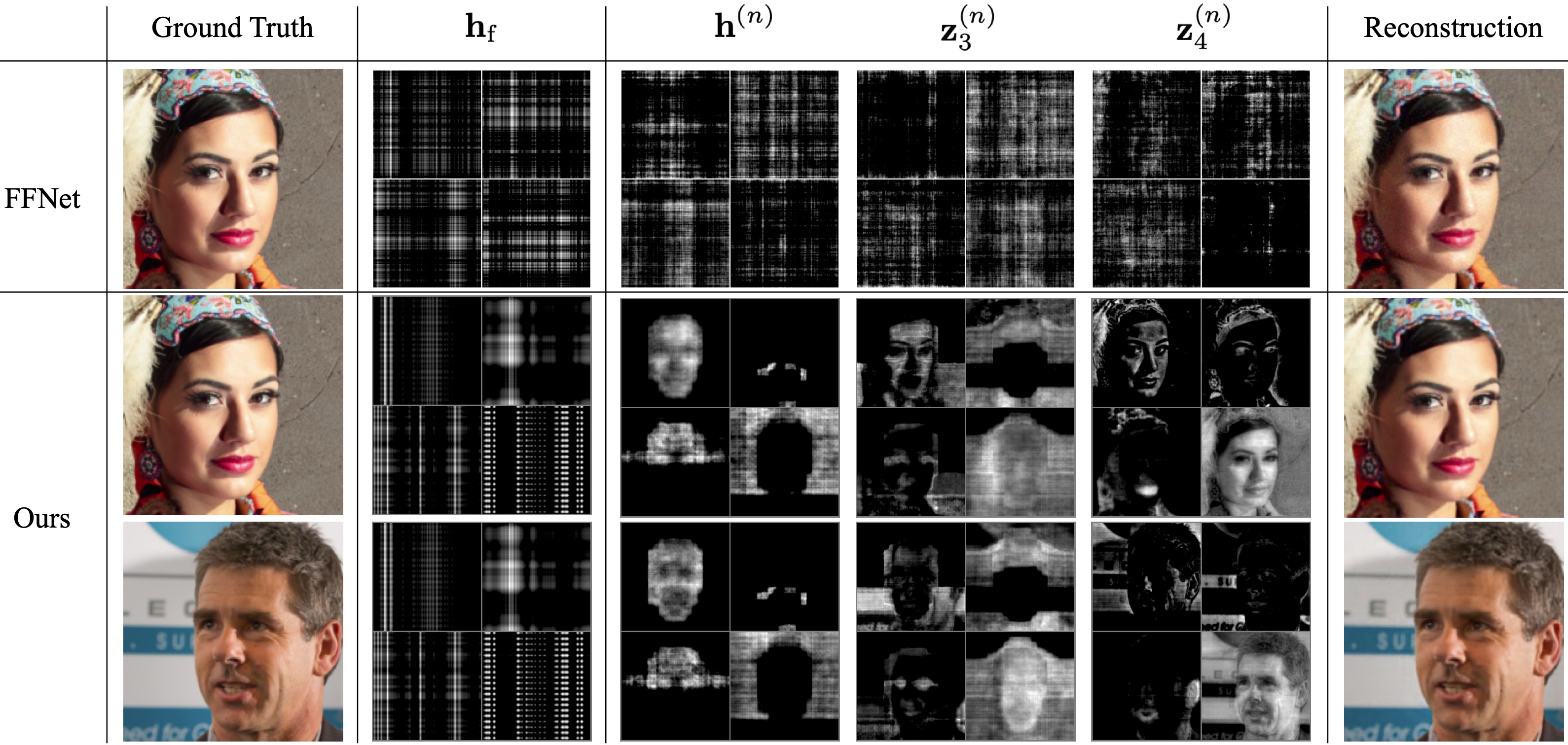}
    \caption{Activation maps of FFNet~\cite{ffnet} to be separately trained to memorize a data instance (top row) and our generalizable INRs (bottom two rows). We select four neurons from each hidden layer to visualize and interpret the activation maps over input coordinates.}
    \label{fig:viz_activation}
    \vspace{-0.1in}
\end{figure*}

\paragraph{Methods for Weight Modulation}
We first validate the design of our modulation method in Eq.~\eqref{eq:W=UV}, where the modulated weights $\bW^{(n)}$ consists of the matrix multiplication of instance-agnostic weight $\bU$ and instance-specific parameter $\bV^{(n)}$.
We compare our method with three variants in Table~\ref{tab:ablation_mod_type} to predict the modulated weights $\bW^{(n)}$:
the direct prediction $\bW^{(n)}=\bV^{(n)}$, Hadamard product $\bU \odot \bV^{(n)}$, and an instance-specific $\mathbf{U}^{(n)}$.
In the case of $\mathbf{U}^{(n)} \mathbf{V}^{(n)}$, our transformer predicts each column vector of $\mathbf{U}^{(n)}$ and row vector of $\mathbf{V}^{(n)}$.
Although the variants provide reasonable results in Table~\ref{tab:ablation_mod_type}, our method shows high performance on both image reconstruction and novel view synthesis.

\paragraph{The Location of Modulated Weights}
We change the location of weight modulation from the first layer to the fifth layer and evaluate its effects. 
Table~\ref{tab:ablation_mod_scope} shows that modulating the second MLP weight achieves the best performance on image reconstruction of both ImageNette and FFHQ.
Interestingly, the performance on ImageNette significantly deteriorates when we modulate the first layer, which takes Fourier features as its input.
Considering the high complexity of ImageNette, a coordinate-based MLP necessitates complex frequency patterns rather than simple and periodic Fourier features to generate instance content patterns of an image.
The PSNRs also deteriorate when we modulate the third or above layers, since the MLP cannot learn the pattern composition rule enough.
Thus, we modulate the second layer for generalizable INRs in experiments.

\paragraph{Optimization-based Meta-Learning}
\label{exp:ablation_meta}

Our instance pattern composer can improve the performance of generalizable INRs with optimization-based meta-learning in Algorithm~\ref{alg:cavia}.
We train a coordinate-based MLP on FFHQ 256$\times$256 for 100 epochs with $\epsilon = 0.01$, $\epsilon'=0.001$, and $N_{\text{inner}}=2$. 
We also train another model using MAML~\cite{maml} to adapt the whole MLP weights for a data instance during 100 epochs with $N_\mathrm{inner}=2$ and $\epsilon=0.001$ and $\epsilon'=0.0003$.
While the model trained with MAML achieves a PSNR of 32.84 after adaptations, our model achieves a higher PSNR of 33.74.
We remark that our model only adapts one weight matrix $\bV^{(n)}$ in the adaptation steps.
The results imply that a coordinate-based MLP can effectively compose instance content patterns to represent unseen data, while exploiting the shared representations across instances.
Figure~\ref{fig:example_cavia} also shows that our generalizable INRs can adapt to unseen instances after two update steps $\bV^{(n)}$.

\subsection{Visualization Analysis of MLP Activations}
\label{exp:viz_analysis}

Figure~\ref{fig:viz_activation} visualize the activations of trained coordinate-based MLP on FFHQ to understand how our generalizable INRs work.
First, we train a coordinate-based MLP on a data instance seperately~\cite{ffnet}, and visualize the activations of selected neurons in each MLP layer.
However, a semantic structure does not exist in the activation maps, since the MLP is trained to memorize a data instance without learning the underlying structures across different instances.

Contrastively, our generalizable INRs have common structures in activation maps of different images.
After non-periodic and instance-agnostic frequency patterns $\vech_\mathrm{f}$ are composed, instance-specific content patterns $\vech^{(n)}$ are generated.
Regardless of data instances, each neuron in the second layer is activated at similar coordinates, but shows different signals of patterns.
Our instance pattern composers learn to assign each neuron in the second layer to different coordinate regions for generating instance-specific patterns.
Then, while subsequent layers use the instance-agnostic rule to compose $\vech^{(n)}$, each neuron has a similar role across instances, enlarges the activated regions, and synthesizes complex and global patterns as the layer goes up.
That is, our generalizable INRs learn underlying structures across instances to represent complex data as the composition of instance-specific low-level patterns.

\section{Conclusion} \label{sec:conclusion}
This study has proposed the framework for generalizable INRs via instance pattern composers, which modulate one weight matrix of the early MLP layer to generalize the learned INRs for unseen data instances.
Thanks to the simplicity, our framework is compatible with both optimization-based meta-learning and hypernetworks to significantly improve the performance of generalizable INRs.
Experimental results demonstrate the broad impacts of the proposed method on various domains and tasks, since our generalizable INRs effectively learn underlying representations across instances.
Our study remains a theoretical analysis of our generalizable INRs as worth exploration.

\section{Acknowledgements}
This work was supported by Institute of Information \& communications Technology Planning \& Evaluation(IITP) grant funded by the Korea government(MSIT) (No.2018-0-01398: Development of a Conversational, Self-tuning DBMS, 35\%; No.2022-0-00113: Sustainable Collaborative Multi-modal Lifelong Learning, 30\%) and the National Research Foundation of Korea(NRF) grant funded by the Korea government(MSIT) (No. NRF-2021R1A2B5B03001551, 35\%)

{\small
\bibliographystyle{ieee_fullname}
\bibliography{egbib}
}

\clearpage

\appendix

\section{Implementation Details}
We describe the implementation details of our framework for generalizable INRs via instance pattern composers.
In all experiments of this study, our transformer-based hypernetwork with 768 hidden dimensions consists of six self-attention blocks with 12 attention heads, where each attention head has 64 dimensions.
We use Adam~\cite{Adam} with $\beta_1 = 0.9$, $\beta_2 = 0.999$, and the constant learning rate of 0.0001 to train our transformers.
The training epochs are different in experiments, and we describe the details below.
Given a rank $r$ of $\bU$ and $\bV^{(n)}$, and the hidden dimension $d$ of the MLP, we initialize $\bU \sim \mathcal{N}(0,1/\sqrt{rd})$ for the stability of training, since we intend to scale the initialization of $\bW_\mathrm{m}^{(n)}$ as $\bW_\mathrm{m}^{(n)} \sim \mathcal{N}(0,1/\sqrt{d})$ while $\bV^{(n)} \sim \mathcal{N}(0, 1)$ at initialization.
We also use weight standardization~\cite{weightstandardization} for the weights of coordinate-based MLPs.

\subsection{Audio Reconstruction}
We train our framework on the train split of LibriSpeech-clean~\cite{librispeech} for the audio reconstruction, while evaluating on its test split.
We use 200 sizes of non-overlapping patches to unfold and tokenize each audio instance, which is sampled by 16kHz, and then a second of audio is expressed as a sequence of 80 data tokens.
Since we train our generalizable INRs to represent one or three seconds of audios, we randomly crop training audio.
For evaluation, we trim a test audio instance into one or three seconds for audio.
We train both our framework and TransINR for 1,000 epochs.
For a fair comparison with TransINR, which predicts 257 weight tokens, our transformer-based hypernetwork predicts $r=256$ weight tokens for instance pattern composers.
A coordinate-based MLP has five layers with $d=256$, where $d_\text{in}=1$ and $d_\text{out}=1$.

\subsection{Image Reconstruction}
We evaluate our generalizable INRs on image reconstruction on facial images such as CelebA~\cite{celeba} and FFHQ\cite{karras2019style}, and natural images of ImageNette~\cite{learnit,transinr}.
We use a zero-padding for 178$\times$178 images to convert them into the 180$\times$180 resolution, and use non-overlapping 9$\times$9 patches to represent an image as the sequence of 400 data tokens.
For 256$\times$256 and 512$\times$512 images, we use 16$\times$16 and 32$\times$32 size of non-overlapping patches, respectively.
A coordinate-based MLP has five layers with $d=256$, where $d_\text{in}=3$ and $d_\text{out}=3$.
We train our framework with $r=256$ and TransINR on 178$\times$178 CelebA, FFHQ, and ImageNette during 300, 1000, and 4000 epochs, respectively, until the training converges.

For 256$\times$256 and 512$\times$512 FFHQ, a model is trained during 400 epochs due to the limited computational resources, but the performance consistently improves as we train the model longer.
In addition, considering the balance of computational costs of transformers and MLP for high-resolution images, we subsample 10\% of coordinates to compute the mean-squared error in Eq.~\eqref{eq:ginr_loss}.

\subsection{Novel View Synthesis}
We evaluate our framework with $r=256$ on novel view synthesis of a 3D object based on the ShapeNet Chairs, Cars, and Lamps datasets.
We follow the experimental settings of previous studies, TransINR~\cite{learnit,transinr}, except for the manual decay of learning rate in TransINR~\cite{transinr}, but use a constant learning rate until the training converges.
We train our framework and TransINR for 1000 epochs for Chairs and Cars until the training converges.
However, we use 400 epochs for Lamps, since TransINR starts overfitting after 400 epochs, although our framework consistently improves the performance.
Before we tokenize each image, we concatenate the starting point and direction of each emitted ray from every pixel into the RGB channels of each pixel, and then each spatial coordinate has nine channels of features.
Given 128$\times$128 resolution of images per view, we use 8$\times$8 non-overlapping patches to represent each image as the sequence of 256 data tokens. 
When multiple support views are used, we concatenate the data tokens of each view as one sequence.
A coordinate-based MLP has six layers with $d=256$, $d_\mathrm{in}=3$, $d_\mathrm{out}=4$.
We use adaptive random sampling~\cite{transinr} during the first epoch to stabilize the training.
We subsample 128 rays during training.

\section{Examples of Novel View Synthesis}
\begin{figure*}
    \centering
    \begin{subfigure}{0.4\textwidth}
    \includegraphics[width=\textwidth]{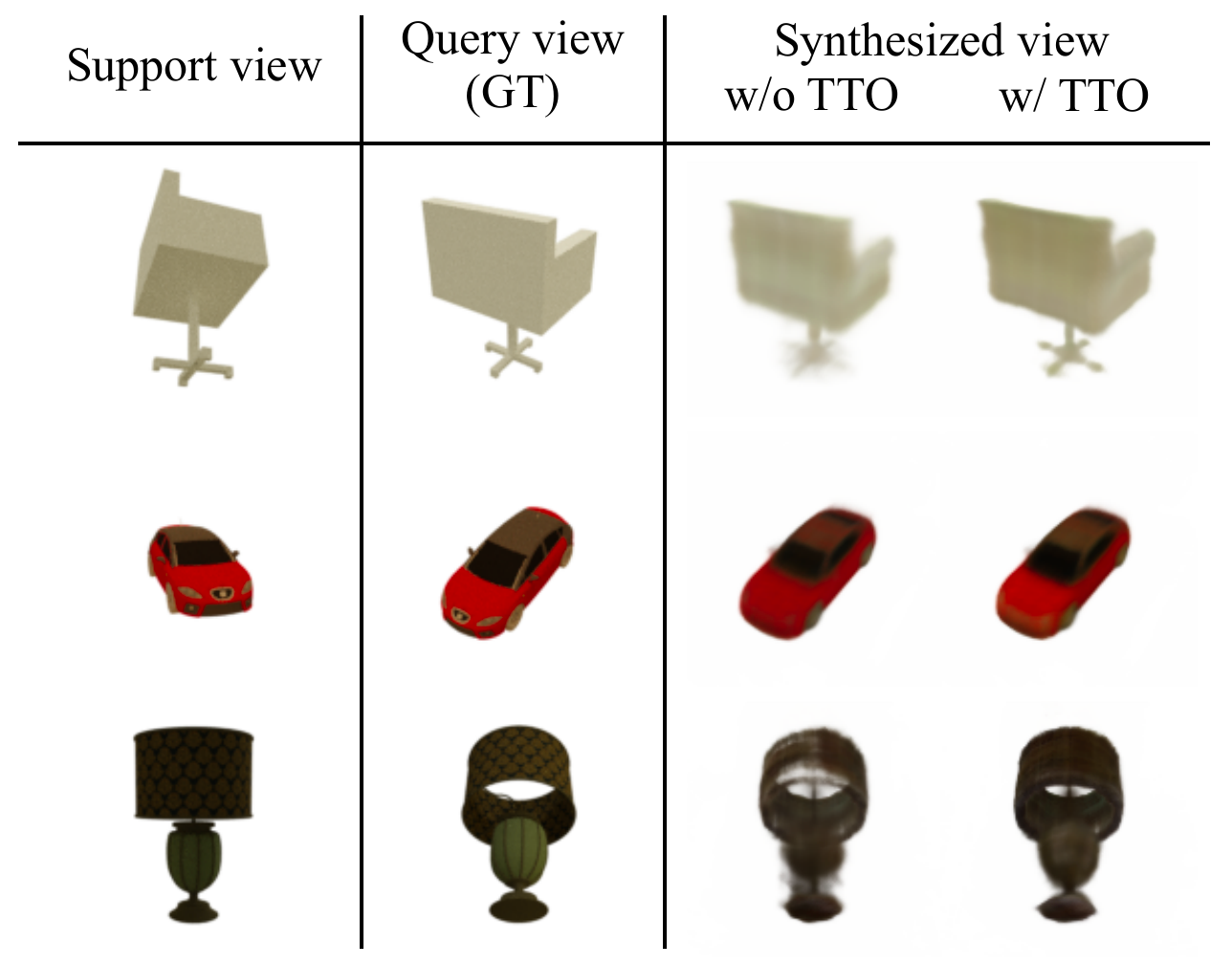}
    \caption{With one support view.}
    \end{subfigure}
    \hspace{0.01\textwidth}
    \begin{subfigure}{0.54\textwidth}
    \includegraphics[width=\textwidth]{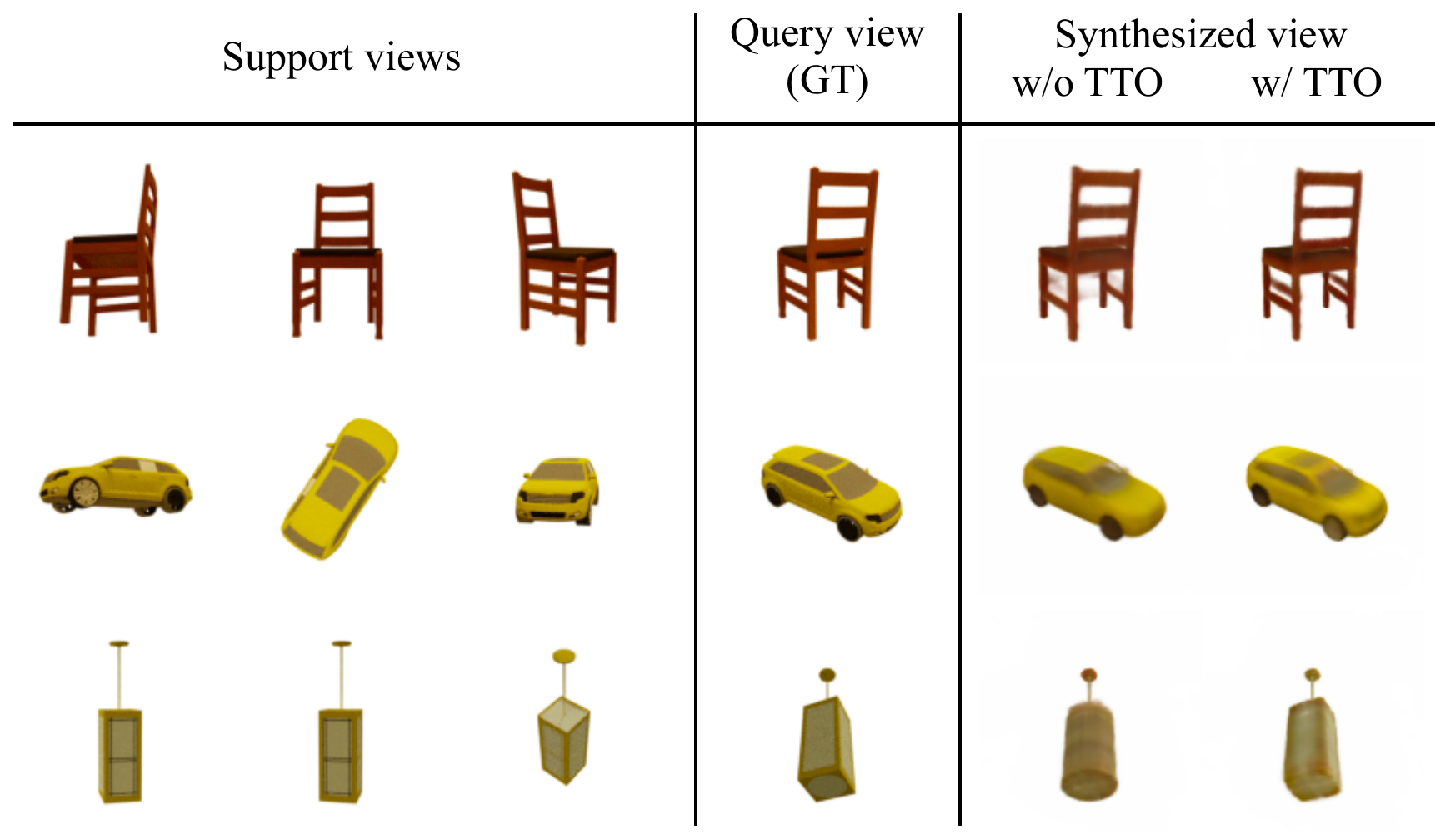}
    \caption{With three support views.}
    \end{subfigure}
    \\
    \vspace{10pt}
    \begin{subfigure}{\textwidth}
    \centering
    \includegraphics[height=0.315\textwidth]{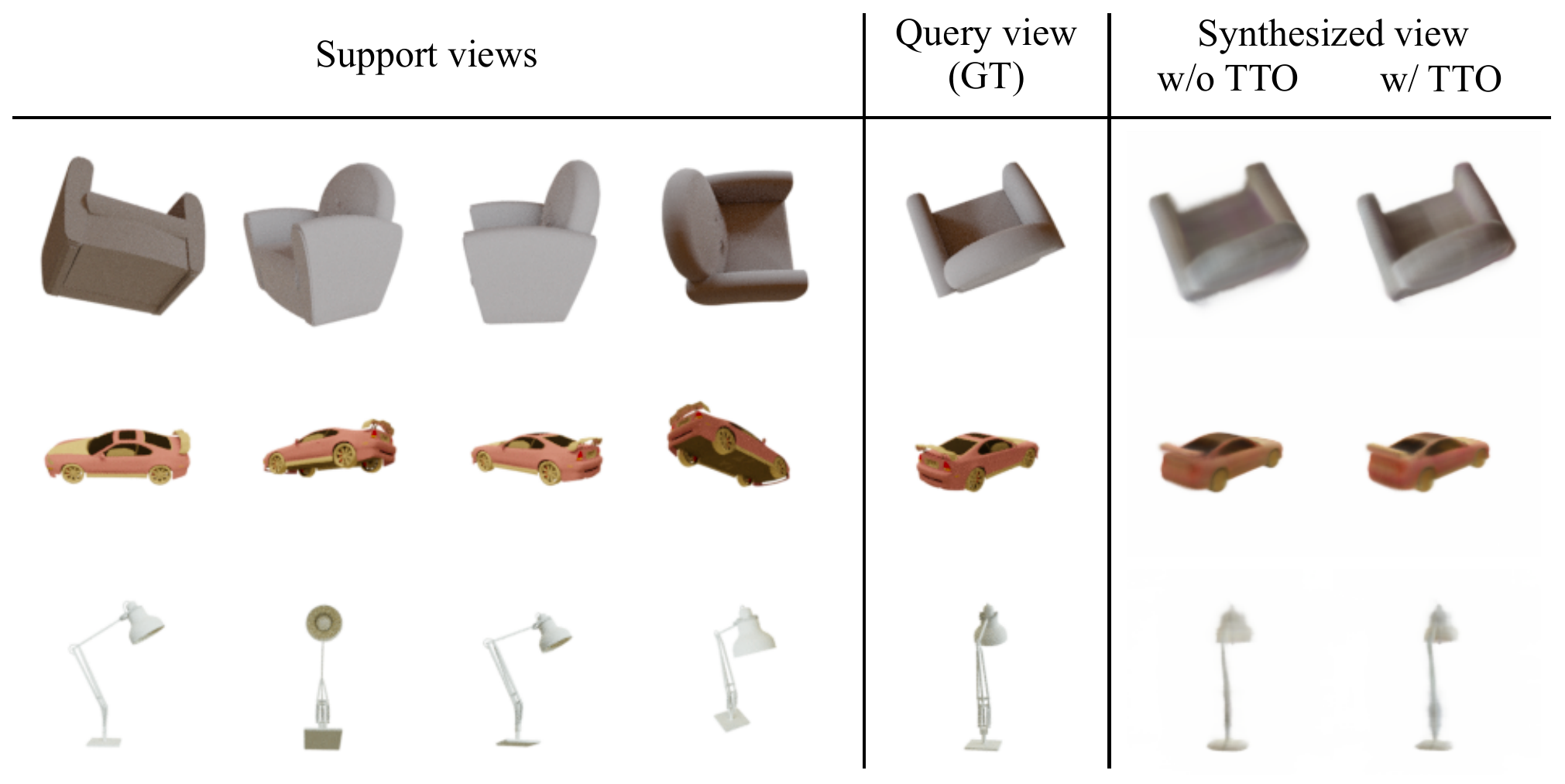}
    \caption{With four support views.}
    \end{subfigure}
    \\
    \vspace{10pt}
    \begin{subfigure}{\textwidth}
    \centering
    \includegraphics[height=0.315\textwidth]{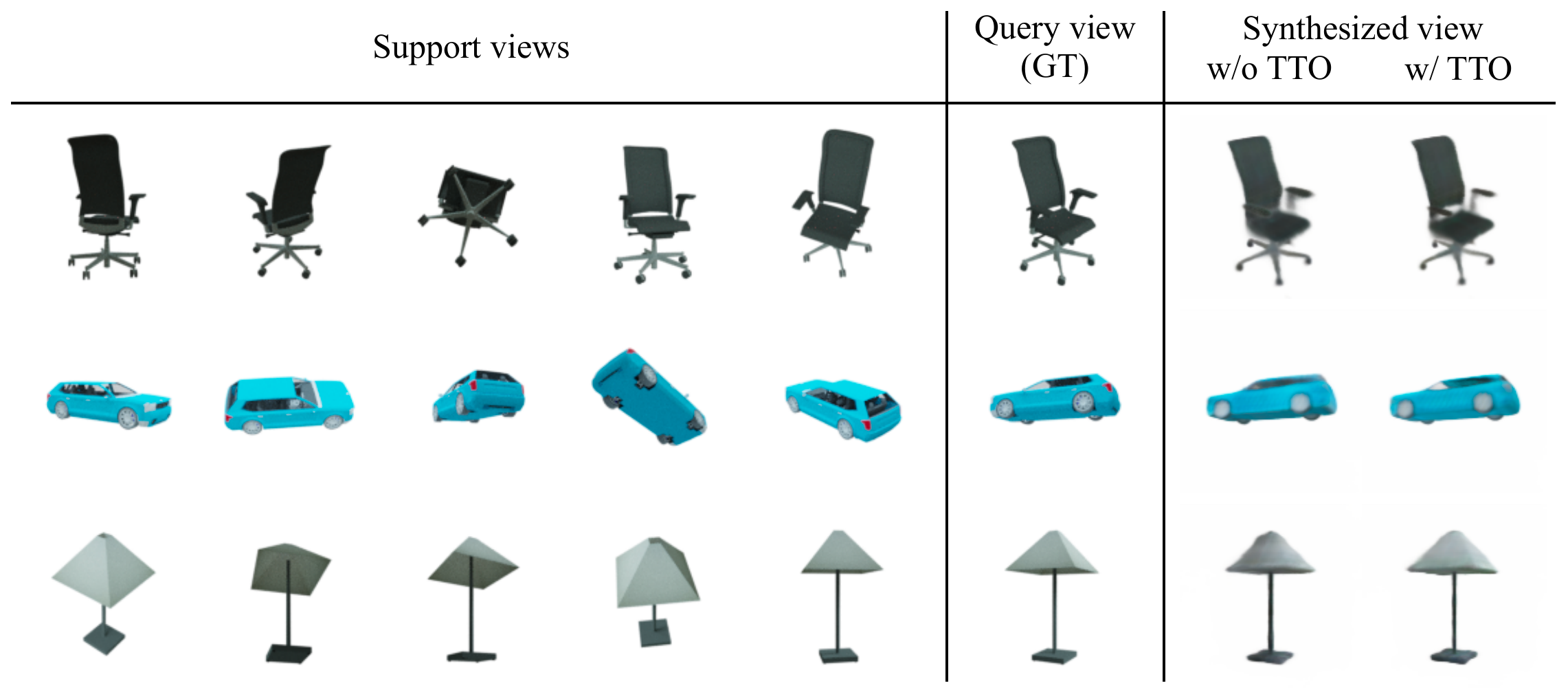}
    \caption{With five support views.}
    \end{subfigure}
    \caption{The examples of Novel view synthesis on Chairs, Cars, and Lamps by our framework with one, three, four, and five support views (a-d).}
    \label{fig:nvs_multiview_examples}
\end{figure*}

Figure \ref{fig:nvs_multiview_examples}, we attach more examples of novel view synthesis on ShapeNet Chairs, Cars, and Lamps by our framework, where the number of support views is increased from one to five.
As shown in Table~\ref{tab:nvs_tto}, the quality of synthesized images increases as the number of support views increases, while our framework modulates only one weight matrix of the instance pattern composer.

\section{Comparison with overfitted INRs} 
\begin{figure}
    \centering
    \includegraphics[width=0.45\textwidth]{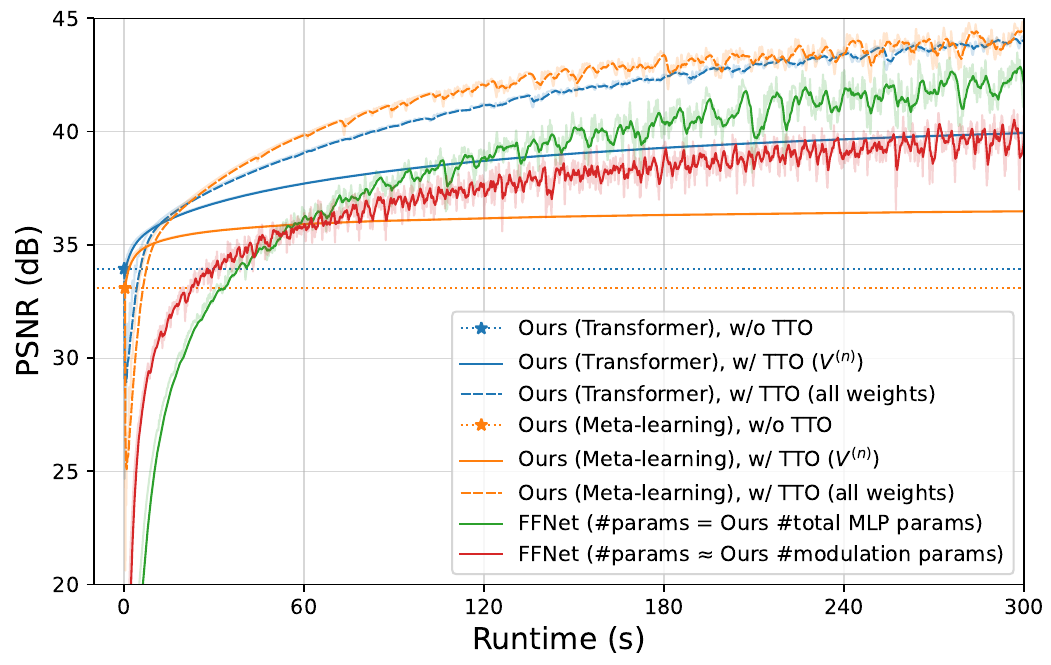}
    \caption{Time/PSNR trade-off during training of randomly initialized FFNets and our generalizable INRs. Each FFNet is trained per a sample in randomly selected 10 images in FFHQ 256$\times$256.}
    \label{fig:rebuttal_time_psnr}
    \vspace{-0.15in}
\end{figure}

Figure~\ref{fig:rebuttal_time_psnr} shows the efficiency to represent data as INRs with or without individual training of MLPs.
Since our framework exploits FFNets for INRs, we perform test-time optimization (TTO) on FFNets initialized 1) randomly, 2) by our meta-learning, and 3) by our transformer-based hypernetwork.
Since the inference time for weight modulation is shorter than one optimization step, the time is negligible, while providing meaningful representations without individual training of FFNets.
When the number of trainable parameters is equivalent, the PSNRs converge to similar values after TTO, but our TTO w/ $\mathbf{V}^{(n)}$ can maintain interpretable structures in Figure~\ref{fig:viz_activation_ours}.
When we optimize the entire weights of INRs, our framework shows better performance during training than random initialization. 

\section{Visualization Analysis of MLP Activations}
\begin{figure*}
    \centering
    \includegraphics[width=0.98\textwidth]{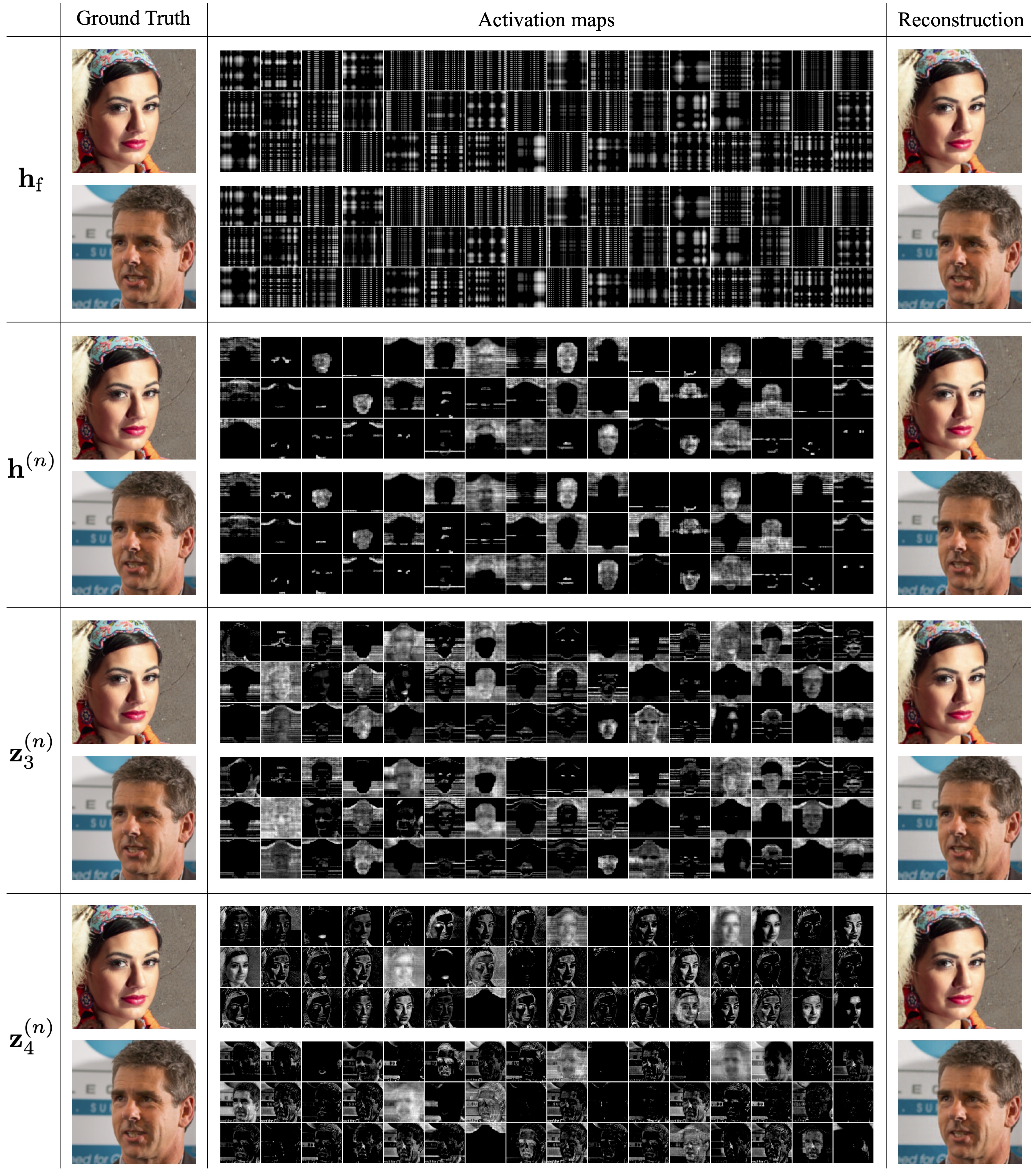}
    \caption{Activation maps of two INRs predicted by our framework of generalizable INRs for each of the two images.}
    \label{fig:viz_activation_ours}
\end{figure*}

After we separately train a coordinate-based MLP, denoted as FFNet~\cite{ffnet}, on each image, we visualize the activation patterns of each neuron in a layer over all coordinate inputs.
Since the two FFNets memorize their training sample separately, the activation maps neither capture the common representations across instances nor be easily interpreted.
On the other hand, Figure \ref{fig:viz_activation_ours} shows that our generalizable INRs can exploit the instance-agnostic pattern composition rule, which enables each neuron to capture common and interpretable structures across instances.
The visualization of activation maps validates that our framework enables the learned representations and pattern composition rule of coordinate-based MLP to be effectively generalized to unseen data instances.

\begin{figure*}
    \centering
    \includegraphics[width=0.98\textwidth]{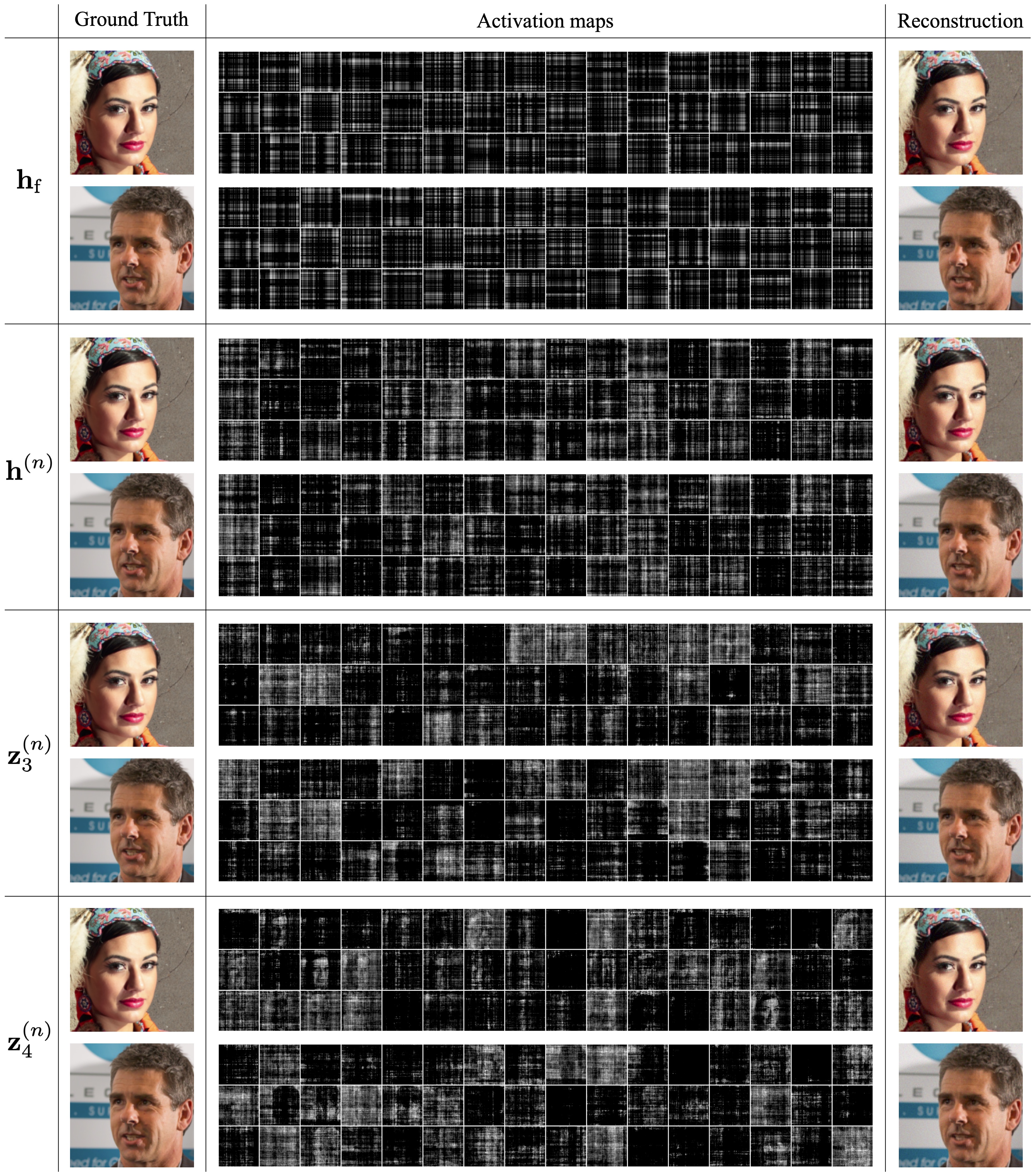}
    \caption{Activation maps of two FFNets~\cite{ffnet}, which are separately trained to memorize each of the two images.}
    \label{fig:viz_activation_ffnet}
\end{figure*}

\end{document}